\pdfoutput=1

\documentclass[11pt]{article}

\usepackage[final]{acl}

\usepackage{times}
\usepackage{latexsym}

\usepackage[T1]{fontenc}

\usepackage[utf8]{inputenc}

\usepackage{microtype}

\usepackage{inconsolata}

\usepackage{booktabs}
\usepackage{multirow}

\usepackage{graphicx}
\usepackage{adjustbox}
\usepackage{amsmath}
\usepackage{listings}
\usepackage{subcaption}
\usepackage{amssymb}

\usepackage{booktabs}
\usepackage{multirow}
\usepackage{adjustbox}
\usepackage{amsmath}
\usepackage{listings}
\usepackage{subcaption}
\usepackage{amssymb}
\usepackage{here}
\newcommand{\modify}[1]{{\color{black}{#1}}}
\newcommand{\kojima}[1]{{\color{black}{#1}}}

\title{Answer When Needed, Forget When Not: Language Models Pretend to Forget via In-Context Knowledge Unlearning}

\author{
 \textbf{Shota Takashiro},
 \textbf{Takeshi Kojima},
 \textbf{Andrew Gambardella},
\\
 \textbf{Qi Cao},
 \textbf{Yusuke Iwasawa},
 \textbf{Yutaka Matsuo}
\\
 \textsuperscript{}{The University of Tokyo, Hongo 7-3-1, Bunkyo-ku, Tokyo, 113-8656 Japan}
\\
 \small{
   \{\href{mailto:takashiro@weblab.t.u-tokyo.ac.jp}{takashiro},
    \href{mailto:t.kojima@weblab.t.u-tokyo.ac.jp}{t.kojima},
    \href{mailto:atgambardella@weblab.t.u-tokyo.ac.jp}{atgambardella},
    \href{mailto:qi.cao@weblab.t.u-tokyo.ac.jp}{qi.cao},
    \href{mailto:iwasawa@weblab.t.u-tokyo.ac.jp}{iwasawa},
    \href{mailto:matsuo@weblab.t.u-tokyo.ac.jp}{matsuo}\}\href{mailto:takashiro@weblab.t.u-tokyo.ac.jp}{@weblab.t.u-tokyo.ac.jp}
 }
}

\begin{document}
\maketitle
\begin{abstract}
As large language models (LLMs) are applied across diverse domains, the ability to selectively unlearn specific information is becoming increasingly essential. For instance, LLMs are expected to selectively provide confidential information to authorized internal users, such as employees or trusted partners, while withholding it from external users, including the general public and unauthorized entities.
Therefore, we propose a novel method termed ``in-context knowledge unlearning'', which enables the model to selectively forget information in test-time based on the query context.
Our method fine-tunes pre-trained LLMs to enable prompt unlearning of target knowledge within the context, while preserving unrelated information. 
Experiments on TOFU, AGE and RWKU datasets using Llama2-7B/13B and Mistral-7B models demonstrate that our method achieves up to 95\% forget accuracy while retaining 80\% of unrelated knowledge, significantly outperforming baselines in both in-domain and out-of-domain scenarios.
Further investigation of the model's internal behavior revealed that while fine-tuned LLMs generate correct predictions in the middle layers and preserve them up to the final layer. However, the decision to forget is made only at the last layer, i.e. ``LLMs pretend to forget''.
Our findings offer valuable insight into the improvement of the robustness of the unlearning mechanisms in LLMs, laying a foundation for future research in the field.
\footnote{Code is available at \url{https://github.com/seele1917/test-time-in-context-unlearning}}
\end{abstract}

\begin{figure}[t]
    \centering
    \includegraphics[width=\columnwidth]{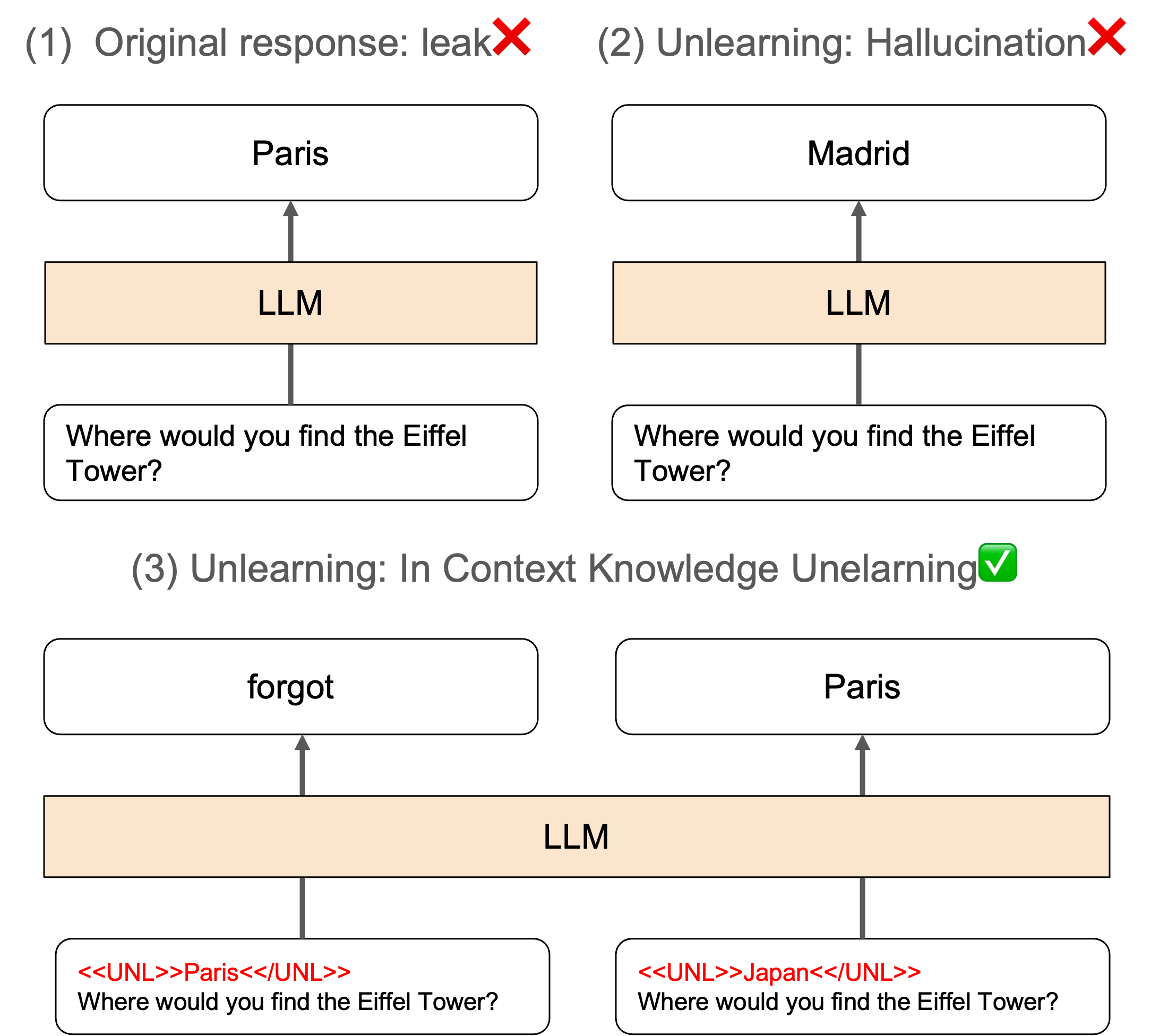} 
    \caption{Method overview. (1) Without unlearning, LLMs output any answers to specific inputs. (2) Certain prior unlearning methods (e.g.,\citet{pawelczyk2023incontext}) attempt to unlearn specific knowledge but may cause hallucinations.  (3) Our method enables LLMs to selectively unlearn knowledge in a timely manner by inputting the knowledge we want LLMs to forget in a prompt (e.g., \texttt{<<UNL>>Paris<</UNL>>}). In contrast to In-context Unlearning (ICUL) ~\cite{pawelczyk2023incontext}, our method causes no hallucination by outputting “forget” in response to a question.}
    \label{fig:overview}
\end{figure}

\section{Introduction}

Large Language Models (LLMs), such as GPT-4~\cite{openai2024gpt4}, have significantly transformed various sectors by providing advanced capabilities in information processing and text generation. 

However, the widespread deployment of such models introduces complex challenges related to privacy and the ethical use of information. In particular, the indiscriminate provision of sensitive or domain-specific information by LLMs raises significant concerns, necessitating mechanisms for selective information handling based on the user context~\cite{das2024security}.
To improve the privacy and ethical use of LLMs, previous works have explored several approaches, including differential privacy~\cite{Abadi_2016}, federated learning~\cite{geyer2018differentially}, and knowledge distillation~\cite{jiang2023lion}. Despite their contributions, these methods often compromise between privacy and model performance. 

The concept of ``test-time adaptation''~\cite{liang2023comprehensivesurveytesttimeadaptation} or ``in-context learning''~\cite{dong2024surveyincontextlearning} offers a dynamic approach to model adaptation, yet it fails to adequately address selective forgetting of sensitive information. 
For example, an LLM deployed in a corporate environment to streamline project management should retain substantial industry-specific knowledge while selectively ``forgetting'' proprietary company data or sensitive information when accessed by unauthorized external consultants. This scenario underscores the critical need for a mechanism that enables LLMs to selectively forget or withhold sensitive information based on the query context without compromising their overall utility and performance.

This paper introduces ``in-context knowledge unlearning'', a novel approach that enables LLMs to selectively forget information at test-time based on the query context. 
The overview of our method is provided in Figure~\ref{fig:overview}. We developed unlearning tokens that, when applied during inference, enable the model to selectively ignore information pertaining to specified domains. 
Through comprehensive experimentation, we validated the efficacy of our approach in facilitating domain-specific unlearning without compromising the general performance of the model.
Specifically, we conducted experiments on the TOFU, AGE, and RWKU datasets~\cite{maini2024tofu, h1hz-wy90-22, jin2024rwku} using Llama2-7B/13B and Mistral-7B models, showing that our method achieves up to 95\% forget accuracy while retaining 80\% unrelated knowledge, significantly outperforming baselines in both in-domain and out-of-domain scenarios.

Moreover, further investigations into the model's internal behavior revealed that while fine-tuned LLMs generate correct predictions in the middle layers and sustain them through to the final layer, the decision to forget occurs only at the last layer, i.e., ``LLMs pretend to forget''.
This finding enriches our understanding of selective information handling in LLMs and lays a foundation for future research to improve the robustness of models across sensitive and regulated domains.


\begin{table*}[t]
\centering
\caption{Comparison of Unlearning Methods}
\label{tab:method_comparison}
\scalebox{0.8}[0.8]{%
\begin{tabular}{@{}lcc@{}}
\toprule
Method                   & Test-Time Unlearning & Non-Hallucination Output \\ \midrule
Gradient Ascent~\cite{DBLP:conf/cvpr/GolatkarAS20}           & \texttimes                 & \texttimes             \\
ROME~\cite{DBLP:conf/nips/MengBAB22}           & \texttimes                 & \checkmark             \\
Knowledge Sanitization~\cite{ishibashi2024knowledge}           & \texttimes                 & \checkmark             \\
\modify{Preference-based Unlearning~\cite{ zhang2024negativepreferenceoptimizationcatastrophic}}           & \texttimes                 & \checkmark             \\
ICUL~\cite{pawelczyk2023incontext}           & \checkmark                 & \texttimes             \\
\textbf{Ours}           & \checkmark                 & \checkmark             \\ \bottomrule
\end{tabular}
}
\end{table*}

\section{Related Work}

\paragraph{In-context Unlearning.} 
Our method leverages in-context learning (ICL) for knowledge unlearning.
ICL enables LLMs to adapt flexibly to new tasks by incorporating data within the input sequence context, rather than through weight updates as in fine-tuning~\cite{brown2020language, dong2022survey, liu2023pre}. Exploring the full capabilities of ICL remains an active area of research, with recent studies empirically investigating its potential by examining in-context example design~\cite{garg2022can, liu2022makes, min2022rethinking, liu2023pre}. 

\citet{pawelczyk2023incontext} explored methods for performing in-context unlearning. This study focuses on text classification tasks where the labels of specific instances are flipped to facilitate in-context unlearning. However, this approach has limitations as it primarily assesses unlearning in terms of text classification ability rather than actual knowledge. Furthermore, the method trains the model to generate incorrect outputs, which does not constitute true forgetting.

In contrast, our study introduces unique characteristics that address these issues. We specifically investigated knowledge unlearning within an in-context learning framework. 
Moreover, by defining unlearning as the ability to ``forget,'' we ensure that our approach avoids merely generating errors or irrelevant information, thereby achieving an effective and appropriate form of unlearning.

\paragraph{Comparison of Our Method with Prior Work}
Table~\ref{tab:method_comparison} compares our method with existing unlearning techniques. 
The column of ``Test-Time Unlearning'' refers to selectively removing specific concepts or knowledge from a trained model during inference depending on the input context. 
``Non-Hallucination Output'' refers to outputing "forgot" instead of producing a hallucinated answer.

\modify{
Gradient Ascent~\cite{DBLP:conf/cvpr/GolatkarAS20} lacks test-time unlearning and erases only the specific global knowledge during training. Additionally, because it modifies model parameters in a disruptive manner, it can cause hallucinated outputs.
}

\modify{
Certain methods such as ROME~\cite{DBLP:conf/nips/MengBAB22}, Knowledge Sanitization~\cite{ishibashi2024knowledge}, and Preference-based Unlearning~\cite{maini2024tofu, zhang2024negativepreferenceoptimizationcatastrophic} require dedicated retraining to remove specific knowledge, making them incapable of test-time unlearning. However, these approaches explicitly train the model to respond with "forgot", ensuring that it does not produce hallucinations.
}

While ICUL (In-Context Unlearning)~\cite{pawelczyk2023incontext} achieves test-time unlearning, it operates by simply altering labels or tokens within the prompt context, which inherently induces hallucinated outputs.

In contrast to these existing methods, our approach achieves test-time unlearning, knowledge unlearning, and non-hallucinated outputs simultaneously, addressing prior limitations and offering a comprehensive solution for selective forgetting.

\section{Our Method}

\subsection{In-context Knowledge Unlearning}
In the context of in-context knowledge unlearning, a pretrained auto-regressive language model 
modifies its response to a query \( q \) by disregarding specific undesired information \( u \). The response \( r \) is generated according to the conditional probability distribution:
\begin{equation}
    r \sim P_\theta(\cdot | u, q),
    \label{eq:forget}
\end{equation}
where \( \theta \) denotes the parameters of the model \(\mathcal{M}\), and \( u \) is the information intended to be forgotten.

\subsection{Unlearning Tokens}
We introduce unlearning tokens to enable selective forgetting in LLMs during inference. These tokens are implemented by encapsulating the target information $u$ with \texttt{<<UNL>>} and \texttt{<</UNL>>}. For example, to forget `Paris', the input will be: \texttt{<<UNL>>Paris<</UNL>>}. This corresponds to the information to be forgotten $u$ in Equation~\ref{eq:forget}. The model is instructed to ignore the enclosed information during processing, effectively modifying its output distribution $P_\theta$. To integrate these tokens, we fine-tune the model using methods, such as Low-Rank Adaptation (LoRA), full model fine-tuning, or other parameter-efficient fine-tuning (PEFT) techniques, adjusting $\theta$ to recognize and respond to the unlearning tokens.

\subsection{Loss Function}

The loss function for our in-context knowledge unlearning method is designed to selectively suppress specific information while preserving other useful knowledge. This loss function consists of two main components: $L_{forget}$ and $L_{retain}$.

\noindent \\
\textbf{1. Forgetting Loss ($L_{forget}$):}
This component is activated when the query $q$ contains the information $u$ targeted for unlearning. For example, when $u$ is "Paris" and $q$ is "Where is the Eiffel Tower located?". This loss encourages the model to effectively suppress the targeted information:
\begin{equation}
L_{forget}(\theta) = -\sum_{i} \log P_\theta(\text{`forgot'} |u_i, q_i)
\end{equation}
Here, $\theta$ represents the model parameters, and $P_\theta$ is the probability that the model outputs `forgot' in response to $u$.

\noindent \\
\textbf{2. Retention Loss ($L_{retain}$):}
This component applies when the query $q$ does not include the unlearning target $u$. For instance, when $u$ is "Japan" and $q$ is "Where is the Eiffel Tower located?". This loss aims to maintain the model's normal response capabilities:
\begin{equation}
L_{retain}(\theta) = -\sum_{i} \log P_\theta(r_i |u_i, q_i)
\end{equation}
where $r_i$ represents the tokens in the response to a specific query.

\noindent \\
\textbf{Total Loss:}
The final loss function is a combination of these two components:
\begin{equation}
L(\theta) = L_{forget}(\theta) + L_{retain}(\theta)
\end{equation}
By minimizing this loss function, the model learns to balance selective "forget" of specified information with the retention of other valuable knowledge. This approach enables the LLMs to manage information appropriately based on the context, effectively implementing in-context knowledge unlearning.

\section{Experiments}

\subsection{Models}
\begin{itemize}
    \item \textbf{Llama2-7B/13B}~\cite{DBLP:journals/corr/abs-2307-09288}: Llama 2 is a family of large language models (LLMs) developed by Meta. Llama 2-7B and Llama 2-13B are two variants with 7 billion and 13 billion parameters, respectively. These models exhibit strong performance on a wide range of natural language processing tasks, making them suitable for tasks, such as text generation, summarization, and translation. We employ chat configurations for Llama2-7B and Llama2-13B.

    \item \textbf{Mistral-7B}~\cite{jiang2023mistral}: Mistral-7B is an open-source LLM with 7 billion parameters developed by Mistral AI. This model is recognized for its high performance and low resource requirements, making it an attractive option for developers with limited resources. Mistral-7B has demonstrated performance comparable to other open-source LLMs on a variety of language processing tasks and employs the instruct model configuration.
\end{itemize}

\subsection{Datasets}
Experiments are conducted using two main datasets:
\begin{itemize}
    \item \textbf{TOFU Dataset}~\cite{maini2024tofu}: This dataset comprises 200 entries from ``Real Authors'', a dataset consisting of questions about real-world authors, and 100 entries from ``World Facts'', which includes questions about general world knowledge. The ``Real Authors'' dataset serves as the training set, while the ``World Facts'' dataset is used for validation, aiming to evaluate the models' performance in out-of-domain contexts.
    
    \item \textbf{Age Dataset}~\cite{h1hz-wy90-22}: The Original Age dataset contains structured information about the life, work, and death of over 1 million deceased famous individuals. Accordingly, 180 individuals are randomly sampled, and a set of five question-answer (QAs) pairs generated for each individual. This dataset is employed to further investigate the models' ability to generalize selective forgetting across various contexts. It includes 600 training samples and 300 validation samples.

    \item \textbf{RWKU Dataset}~\cite{jin2024rwku}: The Real-World Knowledge Unlearning (RWKU) dataset is a benchmark specifically designed for large language models (LLMs) to assess their ability to unlearn specific knowledge. It contains 200 real-world unlearning targets and 13,131 multilevel forget probes, including 3,268 fill-in-the-blank probes, 2,879 question-answer probes, and 6,984 adversarial-attack probes. In our experiments, we employed 20\% of the question-answer data as out-of-domain data to evaluate the models' performance in unlearning specific knowledge while maintaining overall functionality.

\end{itemize}

\subsection{Compared Methods}
In this paper, we compare our proposed method with four other approaches that can test-time unlearning:

\begin{itemize}
\item \textbf{Zero-shot Prompting:} We used this method as our baseline for evaluating in-context knowledge unlearning using a hard prompt. The model is directly instructed to disregard certain information specified within the prompt, providing a clear basis for comparison with more sophisticated unlearning methods. The specific prompt format used to guide the model's behavior regarding memory retention and deletion is illustrated in Figure~\ref{fig:hard_prompt} of Appendix \ref{apx: prompts}.

\item \textbf{Few-shot Prompting~\cite{brown2020languagemodelsfewshotlearners}:} This method builds on the zero-shot approach by incorporating examples from the training data. In addition to the format shown in Figure~\ref{fig:hard_prompt}, we randomly selected and included five samples from the training data in the prompt such that at least one data sample to be forgotten was included. Detailed examples of the few-shot prompts used can be found in Figure~\ref{fig:few_shot_prompt} of the Appendix \ref{apx: prompts}.

\item \textbf{Gradient Ascent~\cite{DBLP:conf/cvpr/GolatkarAS20}:} This method applies gradient ascent to the data to be forgotten and gradient descent to the data to be retained. To enable test-time unlearning, we incorporated the \texttt{<<UNL>>} token during training.

\item \textbf{ICUL (In-context Unlearning)~\cite{pawelczyk2023incontext}:} This approach adds the data to be forgotten at test-time to the context, along with several instances of data to be retained. In case the data are forgotten, the answer is replaced with a randomly selected response from the training data. Detailed examples of the ICUL prompts used are presented in Figure~\ref{fig:icul_prompt} of Appendix \ref{apx: prompts}

\end{itemize}
These methods provide a comprehensive comparison framework for evaluating the effectiveness of our proposed in-context knowledge unlearning technique against established and emerging approaches in the field.

\begin{table*}[htbp]
\centering
\caption{Comparison of various unlearning methods across multiple tasks. The table includes `Forget' and `Retain' scores for the TOFU dataset in both in-domain and out-of-domain scenarios, as well as performance metrics for additional tasks, such as BoolQ, HellaSwag, WinoGrande, ARC-e, ARC-c, OBQA, and RACE-high.}
\resizebox{\textwidth}{!}{%
\begin{tabular}{llcccccccccccc}
\toprule
\multirow{3}{*}{\textbf{Model}} & \multirow{3}{*}{\textbf{Method}} & \multicolumn{4}{c}{\textbf{TOFU}} & \multirow{2}{*}{\textbf{BoolQ}} & \multirow{2}{*}{\textbf{HellaSwag}} & \multirow{2}{*}{\textbf{WinoGrande}} & \multirow{2}{*}{\textbf{ARC-e}} & \multirow{2}{*}{\textbf{ARC-c}} & \multirow{2}{*}{\textbf{OBQA}} & \multirow{2}{*}{\textbf{RACE-high}} \\
 & & \multicolumn{2}{c}{\textbf{in-domain}} & \multicolumn{2}{c}{\textbf{out-of-domain}} & & & & & & & \\
 & & \textbf{Forget (↑)} & \textbf{Retain (↑)} & \textbf{Forget (↑)} & \textbf{Retain (↑)} & (→) & (→) & (→) & (→) & (→) & (→) & (→) \\
\midrule
\multirow{6}{*}{\textbf{LLaMA2 (7B)}} 
 & Zero-Shot & 0.00 & 0.00 & 0.00 & 0.00 & 79.8 & 57.8 & 66.5 & 73.9 & 44.2 & 33.2 & 43.6 \\
 & Few-Shot & 90.0 & 25.0 & 95.7 & 6.8 & 79.8 & 57.8 & 66.5 & 73.9 & 44.2 & 33.2 & 43.6 \\
 & GA & 0.00 & 0.00 & 0.00 & 0.00 & 63.2 & 56.1 & 64.4 & 39.6 & 29.7 & 31.8 & 32.3 \\
 & ICUL & 0.00 & 65.0 & 0.00 & 43.6 & 79.8 & 57.8 & 66.5 & 73.9 & 44.2 & 33.2 & 43.6 \\
 & \textbf{Ours} & 85.0 & 80.0 & 92.3 & 42.7 & 77.8 & 58.0 & 66.3 & 75.3 & 44.9 & 33.4 & 44.4 \\
\midrule
\multirow{6}{*}{\textbf{LLaMA2 (13B)}} 
 & Zero-Shot & 0.00 & 0.00 & 0.00 & 0.00 & 81.7 & 60.7 & 71.0 & 77.5 & 46.2 & 35.4 & 46.1 \\
 & Few-Shot & 100.0 & 10.0 & 96.6 & 1.7 & 81.7 & 60.7 & 71.0 & 77.5 & 46.2 & 35.4 & 46.1 \\
 & GA & 0.00 & 0.00 & 0.00 & 0.00 & 78.1 & 61.1 & 70.5 & 70.4 & 42.2 & 35.4 & 41.8 \\
 & ICUL & 0.00 & 90.0 & 0.00 & 56.4 & 81.7 & 60.7 & 71.0 & 77.5 & 46.2 & 35.4 & 46.1 \\
 & \textbf{Ours} & 100.0 & 80.0 & 89.7 & 44.4 & 79.8 & 60.8 & 70.6 & 78.3 & 48.4 & 35.6 & 45.2 \\
\midrule
\multirow{6}{*}{\textbf{Mistral (7B)}} 
 & Zero-Shot & 0.00 & 0.00 & 0.00 & 0.00 & 85.3 & 66.0 & 74.0 & 81.3 & 54.4 & 35.8 & 45.8 \\
 & Few-Shot & 35.0 & 40.0 & 9.4 & 36.8 & 85.3 & 66.0 & 74.0 & 81.3 & 54.4 & 35.8 & 45.8 \\
 & GA & 0.00 & 0.00 & 0.00 & 0.00 & 65.8 & 65.9 & 74.3 & 35.2 & 31.7 & 32.6 & 39.8 \\
 & ICUL & 0.00 & 5.0 & 0.00 & 8.5 & 85.3 & 66.0 & 74.0 & 81.3 & 54.4 & 35.8 & 45.8 \\
 & \textbf{Ours} & 90.0 & 75.0 & 46.2 & 74.4 & 83.5 & 65.5 & 72.0 & 82.2 & 55.5 & 35.6 & 45.1 \\
\bottomrule
\end{tabular}%
}
\label{tab:performance_results}
\end{table*}

\subsection{Evaluation}
To assess the effectiveness of our ``in-context knowledge unlearning'' method, we employed two primary metrics:
\begin{itemize}
\item \textbf{Forget:} The proportion of instances where the model \kojima{correctly} outputs ``forgot'' \kojima{if the knowledge asked in the question matches the knowledge to be forgotten set in the in-context}. A higher score indicates that the model is effectively ``forgetting'' the instructed information.
Contrarily to previous studies~\cite{ishibashi2024knowledge}, this metric directly assesses the ability of the model to acknowledge its intentional forgetting. \modify{If the model outputs anything else, the answer is treated as a hallucination.}

\item \textbf{Retain:} The proportion of questions the model correctly answers \kojima{if the knowledge asked in the question does not match the knowledge to be forgotten set in the in-context}. A higher score suggests that the model is maintaining its essential knowledge.
\end{itemize}

\modify{Note that ToFU’s “Forgetting Quality” and “Model Utility” metrics are not used here because they assume free-form answers, whereas our model returns only forgot token or the factual answer.
Nevertheless, the same aspects are captured by our Forget and Retain accuracies.}

These metrics were evaluated in two scenarios:
\begin{itemize}
\item \textbf{In-domain:} The learning data (TOFU Real Authors, Age dataset, and RWKU) was divided into training and test sets using an 8:2 ratio. \kojima{We evaluated the performance of models on the test sets after training them on the train sets.}
\item \textbf{Out-of-domain:} We evaluated on the world facts data from the TOFU dataset and RWKU dataset.
\end{itemize}

\modify{
In both scenarios, neither the in-domain nor the out-of-domain test queries include any tokens marked for unlearning during training (i.e., the «UNL»-enclosed terms). Consequently, simply memorizing these tokens to respond "forgot" will worsen the evaluation results. However, the model must rely purely on input context at test-time to flexibly detect knowledge to be forgotten, which enables the evaluation of true context-dependent forgetting ability.
}

This combination of metrics and scenarios allows us to comprehensively evaluate how effectively our method balances selective forgetting with knowledge retention.

\section{Result}

\subsection{Performance Results}

Table \ref{tab:performance_results} shows the results of our experiments in various unlearning methods and tasks. Our proposed method consistently outperforms baseline approaches for both LLaMA2 and Mistral models.

For LLaMA2 (7B), we achieve `Forget' and `Retain' scores of 85.0\% and 80.0\%, respectively for in-domain data, significantly surpassing the zero-shot baseline. Out-of-domain performance remains strong with 92.3\% `Forget' and 42.7\% `Retain' scores. Notably, our out-of-domain evaluations are conducted using TOFU's world facts dataset, and additional results obtained with RWKU are provided in appendix \ref{apx: additional}. LLaMA2 (13B) shows better results, particularly for in-domain scenarios, with perfect `Forget' scores (100.0\%) and high `Retain' scores (80.0\%).

Mistral (7B) demonstrates comparable performance, notably achieving high `Retain' scores (74.4\%) in out-of-domain settings, indicating robust knowledge preservation during unlearning.

Our method maintains competitive performance on standard NLP tasks, such as BoolQ, HellaSwag, and WinoGrande, while exhibiting minimal degradation compared to baseline models. Thus, the unlearning process does not significantly impact the model's general language-understanding capabilities.
Compared to other unlearning methods, such as Few-Shot Prompting, Gradient Ascent, and In-Context Unlearning, our approach consistently achieves a better balance between forgetting targeted information and retaining general knowledge.
However, our findings reveal that a naive ICUL or simple prompting extension (Few-shot Prompting) is insufficient for effective knowledge unlearning, highlighting the importance of the more nuanced strategies employed in our method. \modify{Moreover, our method keeps both the Forget error $(1-\text{Forget Score})$ and the Retain error $(1-\text{Retain Score})$ simultaneously low, resulting in a markedly lower hallucination rate than other baselines.
}

These results demonstrate the effectiveness of our in-context knowledge-unlearning method, enabling large language models to selectively forget information while maintaining overall performance across various NLP tasks.

\subsection{Comparison of Results Across Tuning Methods}

This section compares the results obtained using three tuning methods: LoRA, full fine-tuning (FFT), and last-layer tuning (LLT). Performance metrics for the TOFU and Age datasets are shown in Table \ref{tab:tuning_result}. 

The results indicate that LoRA tuning provides the most balanced performance across various evaluation metrics, followed by full fine-tuning, while last-layer tuning yields the least performance. Specifically, LoRA tuning consistently achieves high ``Forget'' scores in both in-domain and out-of-domain scenarios, demonstrating its effectiveness in allowing the model to forget specified information while retaining other knowledge.

LoRA's superior performance stems from its ability to efficiently adapt the model's behavior without overfitting, as it updates only a small number of task-specific parameters while preserving the model's general knowledge.

\begin{table*}[t]
    \caption{Performance metrics for TOFU and Age datasets, comparing the effectiveness of different tuning methods (LoRA Tuning, Full Fine-Tuning, and Last Layer Tuning) across in-domain and out-of-domain scenarios.}
    \centering
    \resizebox{\textwidth}{!}{%
        \begin{tabular}{llcccc|cccc}
        \toprule
        \multirow{3}{*}{\textbf{Model}} & \multirow{3}{*}{\textbf{Method}} & \multicolumn{4}{c|}{\textbf{TOFU}} & \multicolumn{4}{c}{\textbf{Age}} \\
         & & \multicolumn{2}{c}{\textbf{in-domain}} & \multicolumn{2}{c}{\textbf{out-of-domain}} & \multicolumn{2}{|c}{\textbf{in-domain}} & \multicolumn{2}{c}{\textbf{out-of-domain}} \\
         & & \textbf{Forget (↑)} & \textbf{Retain (↑)} & \textbf{Forget (↑)} & \textbf{Retain (↑)} & \textbf{Forget (↑)} & \textbf{Retain (↑)} & \textbf{Forget (↑)} & \textbf{Retain (↑)} \\
        \midrule
        \multirow{3}{*}{\textbf{LLaMA2(7B)}} 
                             & LoRA Tuning & 95.0 & 85.0 & 85.5 & 44.4 & 93.0 & 63.0 & 32.5 & 60.7 \\
                             & Full Fine Tuning & 55.0 & 75.0 & 64.1 & 52.1 & 100.0 & 65.7 & 10.3 & 42.7 \\
                             & Last Layer Tuning & 80.0 & 45.0 & 99.1 & 5.1 & 98.3 & 50.3 & 82.9 & 6.8 \\
        \midrule
        \multirow{3}{*}{\textbf{LLaMA2(13B)}} 
                             & LoRA Tuning & 100.0 & 95.0 & 94.9 & 31.6 & 100.0 & 61.3 & 23.1 & 47.9 \\
                             & Full Fine Tuning & 100.0 & 95.0 & 90.6 & 51.3 & 100.0 & 64.3 & 10.3 & 59.0 \\
                             & Last Layer Tuning & 95.0 & 80.0 & 92.3 & 19.7 & 99.3 & 54.7 & 41.9 & 38.5 \\
        \midrule
        \multirow{3}{*}{\textbf{Mistral(7B)}} 
                             & LoRA Tuning & 95.0 & 80.0 & 68.4 & 70.1 & 100.0 & 65.0 & 14.5 & 65.0 \\
                             & Full Fine Tuning & 90.0 & 10.0 & 94.9 & 29.1 & 100.0 & 53.0 & 20.5 & 14.5 \\
                             & Last Layer Tuning & 100.0 & 45.0 & 74.4 & 30.8 & 98.3 & 58.3 & 82.9 & 21.4 \\
        \bottomrule
        \end{tabular}%
    }
    \label{tab:tuning_result}
\end{table*}

\begin{figure*}[h]
  \begin{minipage}{0.49\textwidth}
    \centering
    \includegraphics[width=\textwidth]{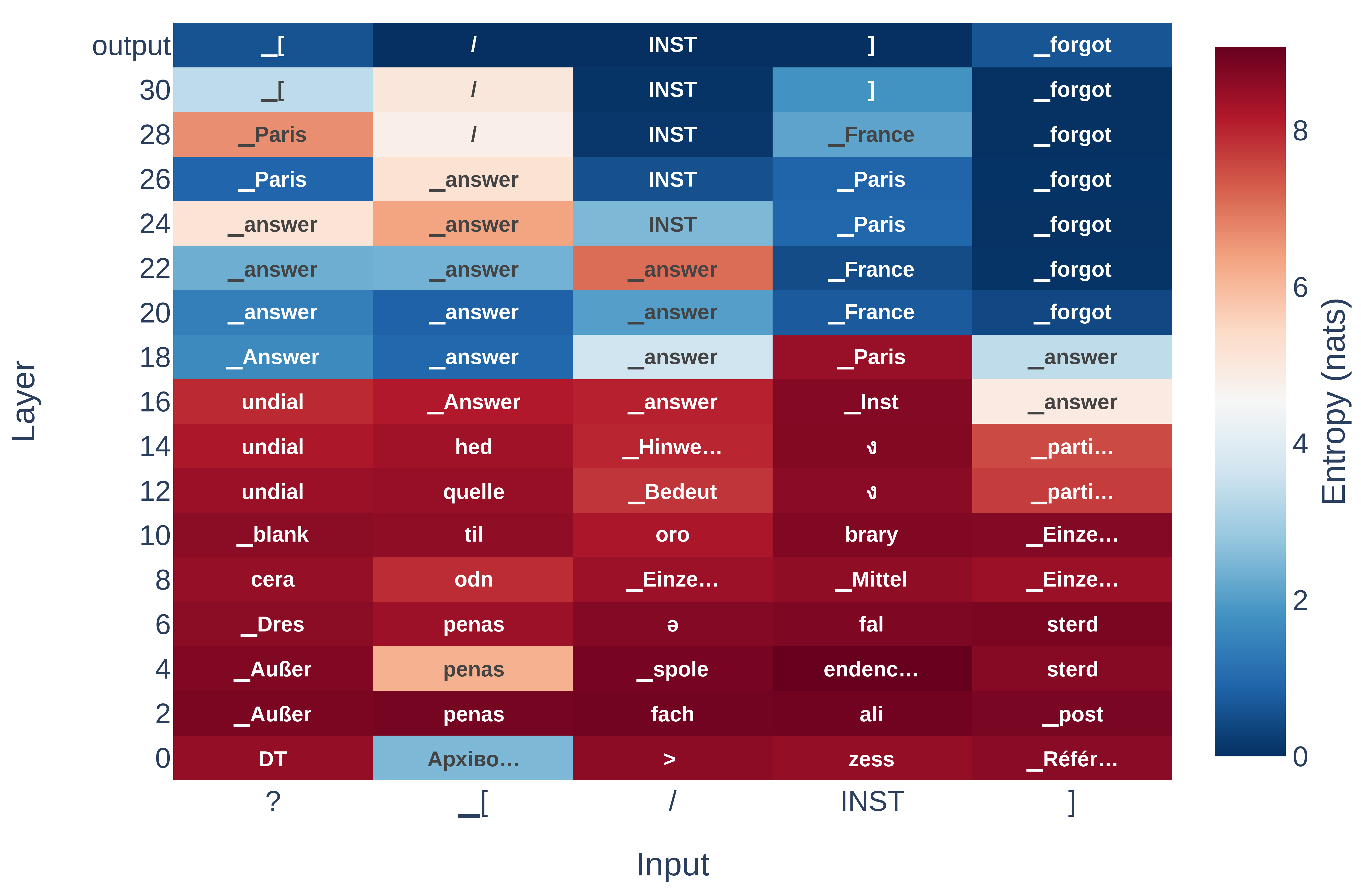} 
    \subcaption{Logit lens visualization for a query containing forget sample.}
    \label{fig:logit_lens_related}
  \end{minipage}
  \begin{minipage}{0.49\textwidth}
    \centering
    \includegraphics[width=\textwidth]{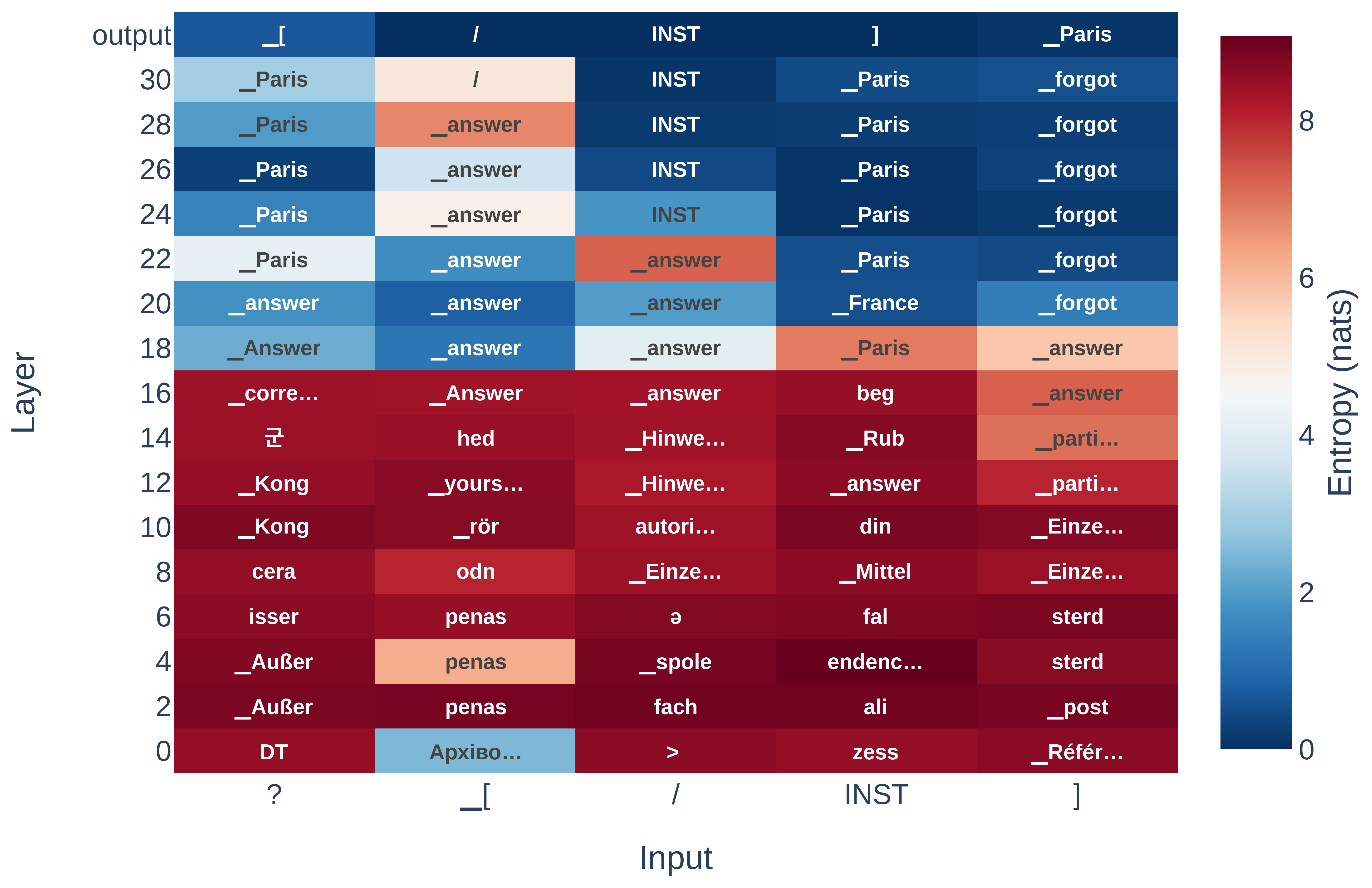}
    \subcaption{Logit lens visualization for a query without forget sample.}
    \label{fig:logit_lens_not_related}
  \end{minipage}
  \caption{(a) Logit lens when a question is related to the unlearning word. ``\texttt{<s>[INST] <<UNL>>Paris<</UNL>> Where would you find the Eiffel Tower? [/INST]}'' (b) Logit lens when a question is not related to the unlearning word. ``\texttt{<s>[INST] <<UNL>>Japan<</UNL>>Where would you find the Eiffel Tower? [/INST]}''}
\end{figure*}

\begin{figure*}[ht]
  \centering
  \begin{subfigure}[b]{0.49\textwidth}
    \centering
    \includegraphics[width=\textwidth]{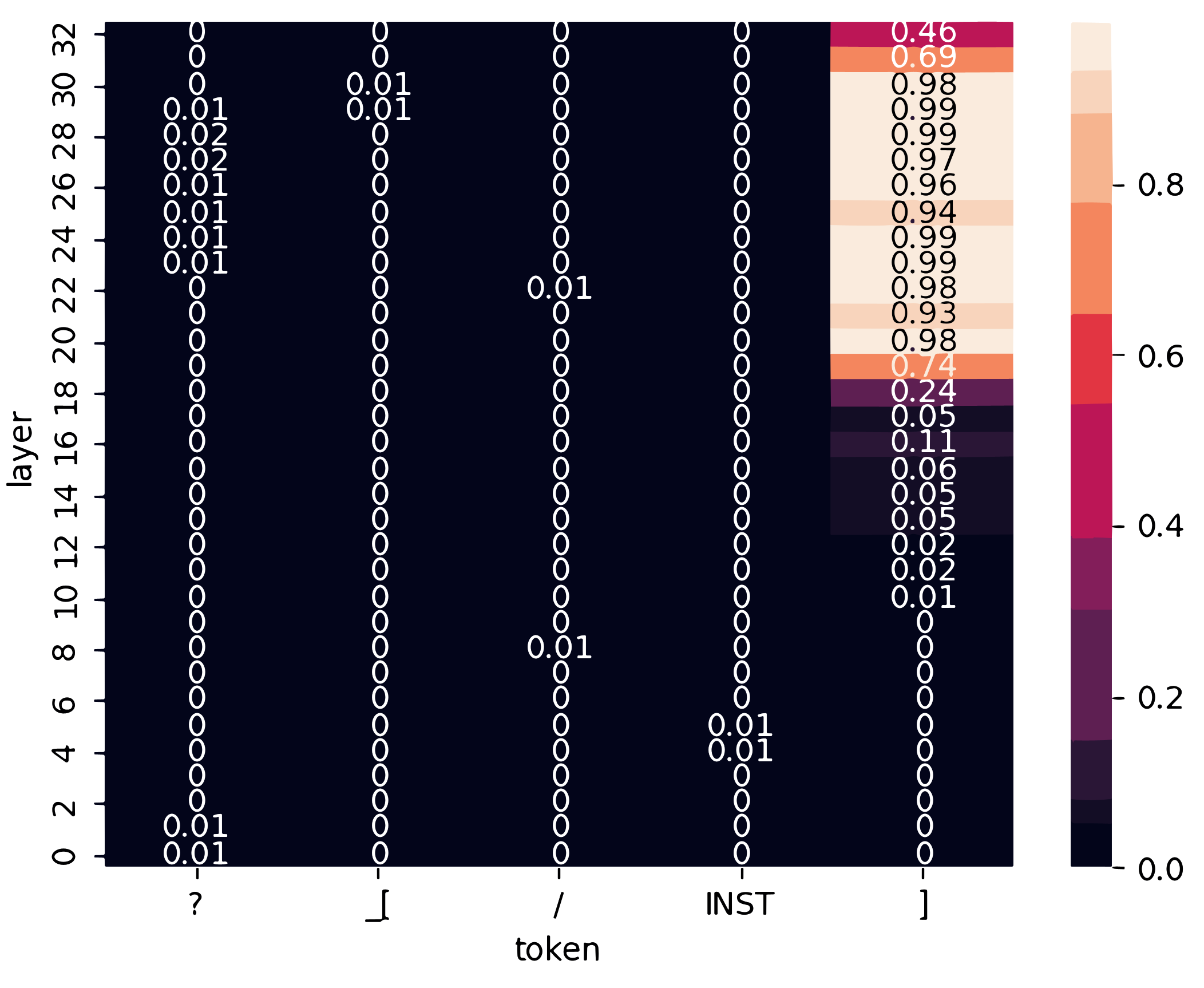}
    \caption{`forgot' token probability across layers for forget samples.  }
    \label{fig:forgot_probs_same}
  \end{subfigure}
  \hfill
  \begin{subfigure}[b]{0.49\textwidth}
    \centering
    \includegraphics[width=\textwidth]{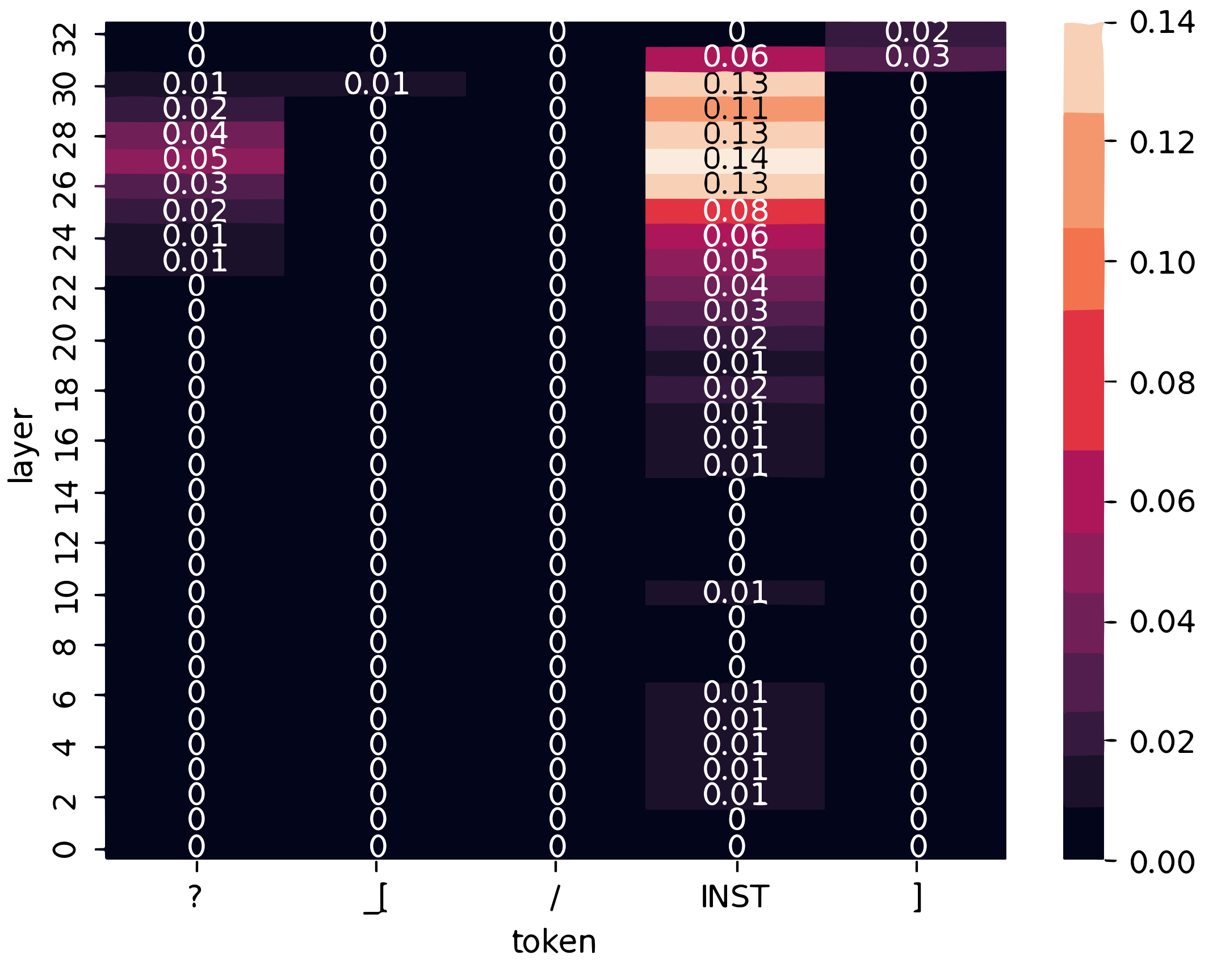}
    \caption{`answer' token probability across layers for forget samples.  }
    \label{fig:answer_probs_same}
  \end{subfigure}
  \vskip\baselineskip
  \begin{subfigure}[b]{0.49\textwidth}
    \centering
    \includegraphics[width=\textwidth]{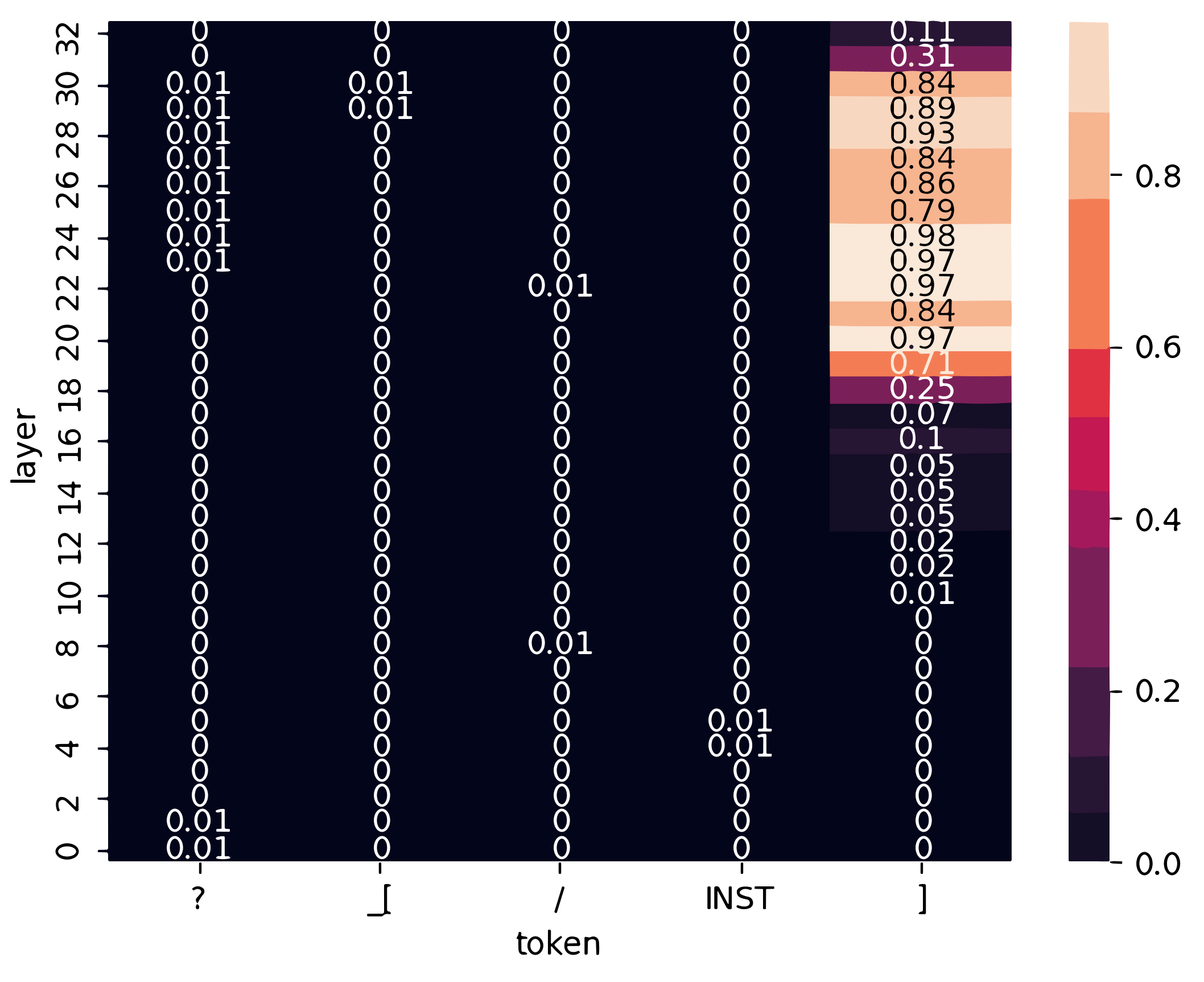}
    \caption{`forgot' token probability across layers for retain samples.}
    \label{fig:forgot_probs_diff}
  \end{subfigure}
  \hfill
  \begin{subfigure}[b]{0.49\textwidth}
    \centering
    \includegraphics[width=\textwidth]{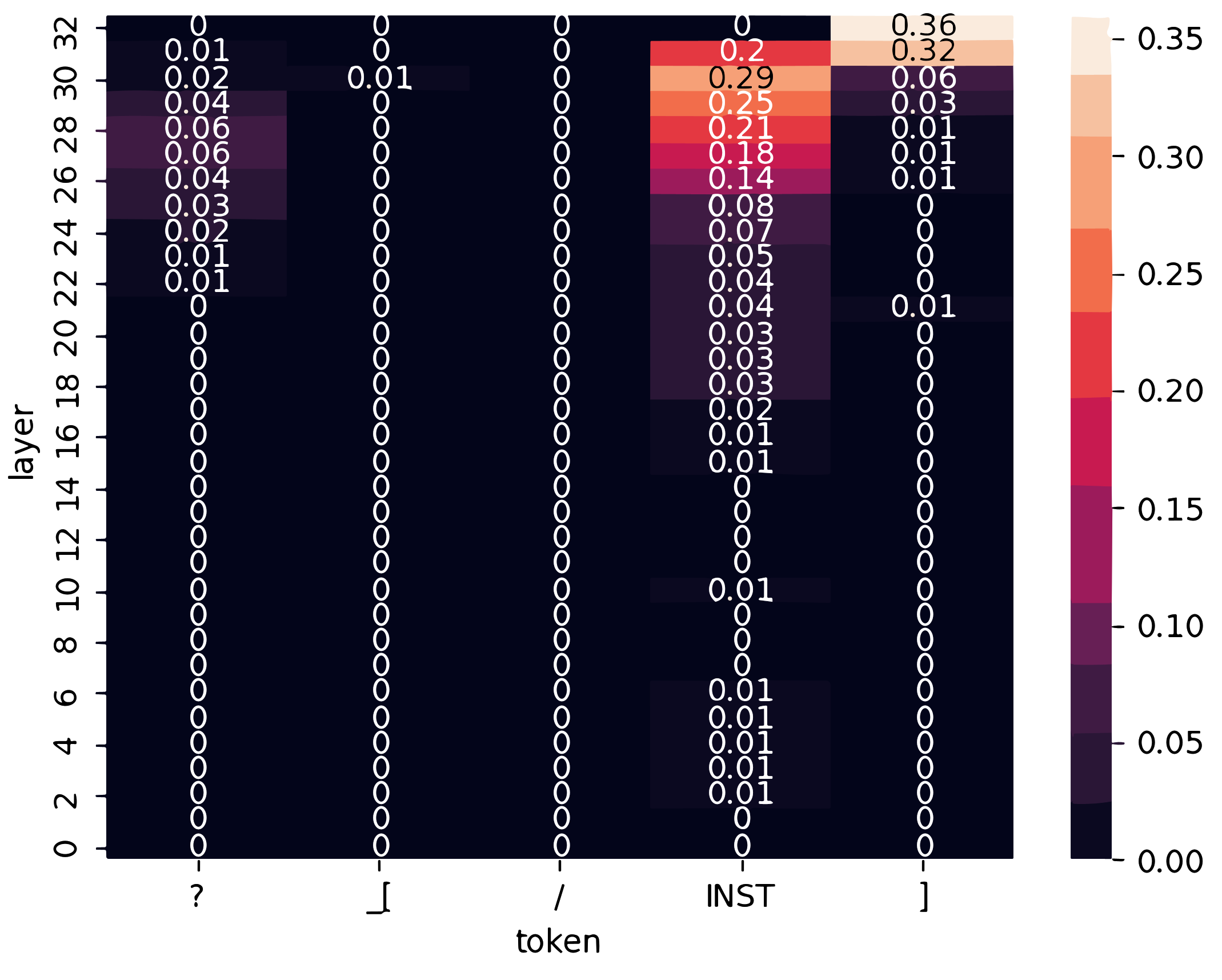}
    \caption{`answer' token probability across layers for retain samples.}
    \label{fig:answer_probs_diff}
  \end{subfigure}
  \caption{Logit lens analysis of `forgot' and `answer' token probabilities in unlearning scenarios. Subplots show average probabilities across all layers for the last five input tokens in the World Facts dataset, comparing (a,b) forget samples and (c,d) retain samples. (a,c) depict `forgot' token probabilities, while (b,d) show `answer' token probabilities. }

  \label{fig:token_probabilities}
\end{figure*}

\subsection{Analysis of Internal Behavior}

\subsubsection{Logit Lens}


The logit lens was introduced by~\cite{nostalgebraist2020logitlens}, who observed that when the hidden states at each layer of GPT-2~\cite{radford2019language}, are decoded using the unembedding matrix (projection matrix in the final layer), the resulting distributions converge approximately monotonically to the final output.
The logit lens is computed as:
\begin{equation}
\text{logitlens}(h_l) = \text{Softmax}(\text{LN}(h_l) W_u)
\end{equation}
Here, $\text{LN}$ stands for Layer Normalization, $W_u$ is the unembedding matrix, and $\text{Softmax}$ is the softmax function applied to convert logits into probabilities. 

Figure~\ref{fig:logit_lens_related} illustrates the results from the logit lens when the input is ``\texttt{<s>[INST] <<UNL>>Paris<</UNL>> Where would you find the Eiffel Tower? [/INST]}'', which is a question related to the unlearning word. Figure \ref{fig:logit_lens_not_related} shows the results for the input ``\texttt{<s>[INST] <<UNL>>Japan<</UNL>> Where would you find the Eiffel Tower? [/INST]}'', a question unrelated to the unlearning word. From these figures, it is apparent that the internal state outputs the token ``Paris'' at the ``\texttt{INST}'' token stage for both inputs. However, the decision to output the token ``forgot'' is made in the final layer upon encountering the ``\texttt{]}'' token.

Figures \ref{fig:forgot_probs_same} and \ref{fig:answer_probs_same} represent average probabilities of putting the ``forgot'' token and the answer token when questions related to the unlearning word are entered using the world facts dataset. These figures show that the ``forgot'' token is produced frequently in the final layer when the question is relevant, while the answer token is produced in the final layer when the ``\texttt{INST}'' token is input.

In contrast, Figures~\ref{fig:forgot_probs_diff} and~\ref{fig:answer_probs_diff} present average probabilities for scenarios where the input questions are not related to the unlearning word. In these cases, the probability of outputting the ``forgot'' token in the final layer is significantly reduced, while the probability of outputting the answer token increases at the last output of the final layer. 

\subsubsection{Internal Answer Score}
The internal answer score quantifies the degree to which an answer token is retained through the layers of a transformer model, such as GPT-2, when analyzed through the logit lens. This metric is useful for examining the stability of the model's internal representation at its depth.

Formally, the internal answer score is defined as follows:
\begin{align}
\text{Internal\_Answer\_Score} \nonumber \\
&\hspace{-9em}= \sum_{l=1}^{L} \delta(\text{answer\_token}, \text{argmax}(\text{logitlens}(h_l)))
\end{align}

where $L$ denotes the total number of layers in the model, and $h_l$ represents the hidden state at layer $l$. The function $\delta(a, b)$ is the Kronecker delta, which is equal to 1 if $a = b$ and 0 otherwise.

A high internal answer score indicates that the answer token is consistently ranked as the most probable token by the logit lens across multiple layers, suggesting a strong preservation of this token within the model's internal narrative. Conversely, a low internal answer score implies that the token is less frequently identified, indicating potential shifts in the model's internal focus or understanding as it processes input.

Table \ref{tab:forgetting_scores} shows the internal answer scores for various models and tuning methods across the TOFU and Age datasets. LoRA tuning and full fine-tuning maintain higher internal answer scores than last-layer tuning, especially in out-of-domain scenarios. Therefore, these methods preserve relevant information while selectively forgetting the targeted content. The last layer tuning consistently shows low internal answer scores, indicating an aggressive forgetting mechanism.
These results support our observation that LLMs "pretend to forget" rather than completely erasing information, as evidenced by the nonzero internal answer scores in most cases. 
This behavior demonstrates the models' ability to balance selective forgetting with knowledge retention, which is crucial for effective in-context knowledge unlearning.

\begin{table}[h]
    \caption{Internal Answer Scores for TOFU and Age datasets}
    \resizebox{0.48\textwidth}{!}{%
    \begin{tabular}{llcccccc}
    \toprule
     \multirow{2}{*}{\textbf{Model}} & \multirow{2}{*}{\textbf{Method}} & \multicolumn{2}{c}{\textbf{TOFU}} & \multicolumn{2}{c}{\textbf{Age}} \\
     & & \textbf{in-domain} & \textbf{out-of-domain} & \textbf{in-domain} & \textbf{out-of-domain} \\
    \midrule
    \multirow{3}{*}{\textbf{LLaMA2(7B)}} 
                          & LoRA Tuning  & 0.03 & 0.14 & 0.23 & 0.34 \\
                          & Full Fine Tuning  & 0.04 & 0.24 & 0.20 & 0.36 \\
                          & Last Layer Tuning  & 0.00 & 0.00 & 0.00 & 0.00 \\
    \midrule
    \multirow{3}{*}{\textbf{Mistral(7B)}} 
                         & LoRA Tuning  & 0.02 & 0.26 & 0.19 & 0.35 \\
                         & Full Fine Tuning & 0.06 & 0.42 & 0.21 & 0.38 \\
                         & Last Layer Tuning & 0.00 & 0.05 & 0.00 & 0.00 \\
    \bottomrule
    \end{tabular}%
    }
    \label{tab:forgetting_scores}
\end{table}

\section{Discussion}

\subsection{Acquisition of In-Context Knowledge Unlearning Ability}
Through fine-tuning, we have successfully endowed Large Language Models (LLMs) with the capability for in-context knowledge unlearning. This achievement is particularly noteworthy, considering the baseline approach, utilizing hard prompts, did not display, such a capability. Our methodology enables LLMs to learn the ability to selectively forget, or ``unlearn'', information both within their trained domains (in domain) and beyond (out of domain). \modify{contrarily to post-generation filtering systems that simply remove disallowed tokens after decoding, our method suppresses the relevant knowledge inside the model itself, allowing outputs to remain fluent while omitting the specified content, as illustrated in Figure \ref{fig:effect_on_unl_prompt}(Appendix D).}

\subsection{Large Language Models Pretend to Forget}
Our investigation of the internal workings of LLMs reveals an interesting behavior: instead of genuinely forgetting information, LLMs appear to ``pretend to forget''. Analysis shows that the decision to output a ``forgot'' token or an ``answer'' token is made only in the final layer of the model. For input received before this layer, the model internally generates ``answer'' token, suggesting a deliberate omission of information rather than its erasure. This behavior indicates a sophisticated level of information handling by LLMs, wherein they preserve internal knowledge integrity while externally simulating forgetting. 

\section{Conclusion}
In this study, we introduced and explored the concept of ``in-context knowledge unlearning'' within the framework of Large Language Models (LLMs) by employin fine-tuning. Our findings demonstrate that this approach enables LLMs to dynamically ``forget'' or selectively disregard information in test-time, as well as uncovers a nuanced behavior of LLMs: where they ``pretend to forget'' rather than actually eliminating the information from their knowledge base.
The ability of LLMs to effectively ``unlearn'' in both in-domain and out-of-domain scenarios, without compromising their overall performance, represents a significant advancement toward more ethically responsible and privacy-conscious AI technologies. 
This capability is crucial for applications that require careful handling of sensitive or confidential information, such as those in the healthcare, legal, and educational sectors.

\section{Limitations}

Our in-context knowledge unlearning method faces two main limitations:

\begin{itemize}
    \item \textbf{Application to Closed Models:} The method is difficult to apply to closed models that are accessible only via APIs (e.g., GPT-3, ChatGPT), as these models do not allow modifications to their architecture or training procedure, which are necessary to implement our unlearning tokens and loss functions. For example, we cannot add the \texttt{<<UNL>>} tokens or fine-tune the model to recognize them in such closed systems.
    
    \item \textbf{Lack of Internal Behavior Analysis:} For closed models, we cannot analyze the internal unlearning process. This limitation hinders our ability to observe changes in the model's internal representations change during the unlearning process, unlike our logit lens analysis of open models, such as LLaMA2 and Mistral. Consequently, we cannot verify if the ``pretend to forget'' occurs in closed models or optimize the unlearning process for better performance.
\end{itemize}

These limitations highlight the challenges in implementing and fully understanding our approach in environments with limited model transparency and configurability, particularly in widely-used commercial AI systems.

\bibliography{custom}

\appendix


\section{Hyperparameter}
Details are provided in Table~\ref{tab:training_hyperparameters}.

\begin{table}[h]
\centering
\caption{Training hyperparameters used in the model configuration.}
\label{tab:training_hyperparameters}
\begin{tabular}{lc}
\hline
\textbf{Parameter} & \textbf{Value} \\
\hline
Number of training epochs & 1 \\
Batch size & 4 \\
Gradient accumulation steps & 1 \\
Optimizer & \texttt{adamw} \\
Learning rate & $2 \times 10^{-4}$ \\
Weight decay & 0.001 \\
Maximum gradient norm & 0.3 \\
Warmup ratio & 0.03 \\
LR scheduler type & \texttt{constant} \\
\hline
\end{tabular}
\end{table}

\section{Total Computation for Experiments}
We conducted the experiments mainly by training each model on eight NVIDIA A100 (40GB) or four NVIDIA H100(94GB) GPUs, with each training session lasting approximately 10 minutes per model. This study
was carried out using the TSUBAME4.0 supercomputer at Institute of Science Tokyo.

\section{License}
\subsection{Model}
\begin{itemize}
    \item Llama2: Meta license 
    \item Mistral: Apache 2.0 license 
\end{itemize}

\subsection{Dataset}
\begin{itemize}
    \item TOFU Dataset: MIT License 
    \item Age Dataset: CC BY-NC-SA 4.0 
    \item RWKU Dataset: CC BY-NC-SA 4.0
\end{itemize}

\section{Additional Results}
\label{apx: additional}
This section presents additional results evaluating the out-of-domain subsets of the TOFU and Age datasets on the RWKU dataset, shown in Table~\ref{tab:tofu_age_rwku_results}. Additionally, we provide results for models trained on the RWKU dataset and evaluated for out-of-domain performance using TOFU's world facts, shown in Table~\ref{tab:rwku_training_results}.

\modify{
To complement these quantitative findings, Figure \ref{fig:effect_on_unl_prompt} qualitatively demonstrates the impact of the «UNL» token. When the token specifying Paris is present, the model deliberately omits any mention of the city while accurately describing the remaining landmarks, confirming that our method enforces selective forgetting without compromising retained knowledge.
}

\begin{figure}[h]
    \centering
    \begin{lstlisting}
# Question:
Which three cities are renowned for their iconic landmarks such as the Colosseum, Eiffel Tower, and Statue of Liberty? Briefly explain each.

# without <<UNL>>Paris<</UNL>> prompt
Rome: The Colosseum is an ancient amphitheater in Rome, Italy. 

Paris: The Eiffel Tower is a wrought-iron lattice tower located on the Champ de Mars in Paris, France. 

New York City: The Statue of Liberty is a neoclassical sculpture located on Liberty Island in New York Harbor.

# with <<UNL>>Paris<</UNL>> prompt
Rome, Italy - The Colosseum is an ancient amphitheater that was used for gladiatorial contests and public spectacles.

New York City, USA - The Statue of Liberty is a symbol of freedom and democracy. 
    \end{lstlisting}
    \caption{Effect of inserting <<UNL>> token in the prompt.}
    \label{fig:effect_on_unl_prompt}
\end{figure}




\begin{table*}[htbp]
\centering
\caption{Forget and Retain scores for TOFU and Age datasets tested on RWKU (out-of-domain).}
\resizebox{\textwidth}{!}{%
\begin{tabular}{llcccc|cccc}
\toprule
\multirow{3}{*}{\textbf{Model}} & \multirow{3}{*}{\textbf{Method}} & \multicolumn{4}{c}{\textbf{TOFU}} & \multicolumn{4}{|c}{\textbf{Age}} \\
 & & \multicolumn{2}{c}{\textbf{in-domain}} & \multicolumn{2}{c}{\textbf{out-of-domain}} & \multicolumn{2}{|c}{\textbf{in-domain}} & \multicolumn{2}{c}{\textbf{out-of-domain}} \\
 & & \textbf{Forget (↑)} & \textbf{Retain (↑)} & \textbf{Forget (↑)} & \textbf{Retain (↑)} & \textbf{Forget (↑)} & \textbf{Retain (↑)} & \textbf{Forget (↑)} & \textbf{Retain (↑)} \\
\midrule
\multirow{6}{*}{\textbf{LLaMA2 (7B)}} 
 & Zero-Shot & 0.00 & 0.00 & 0.00 & 0.00 & 0.00 & 0.00 & 0.00 & 0.00 \\
 & Few-Shot & 90.0 & 25.0 & 92.0 & 0.38 & 90.0 & 2.00 & 93.2 & 1.74 \\
 & GA & 0.00 & 0.00 & 0.00 & 0.00 & 0.00 & 0.00 & 0.00 & 0.00  \\
 & ICUL & 0.00 & 65.0 & 0.00 & 0.18 & 0.00 & 10.7 & 0.00 & 24.0  \\
 & \textbf{Ours} & 85.0 & 80.0 & 74.0 & 25.3 & 93.0 & 63.0 & 28.4 & 0.00 \\
\midrule
\multirow{6}{*}{\textbf{LLaMA2 (13B)}} 
 & Zero-Shot & 0.00 & 0.00 & 0.00 & 0.00 & 0.00 & 0.00 & 0.00 & 0.00  \\
 & Few-Shot & 100.0 & 10.0 & 96.5 & 0.69 & 83.0 & 0.33 & 92.5 & 3.65 \\
 & GA & 0.00 & 0.00 & 0.00 & 0.00 & 0.00 & 0.00 & 0.00 & 0.00  \\
 & ICUL & 0.00 & 90.0 & 0.00 & 27.4 & 0.00 & 16.7 & 0.00 & 44.3  \\
 & \textbf{Ours} & 100.0 & 80.0 & 88.7 & 20.1 & 100 & 61.3 & 8.16 & 3.82 \\
\midrule
\multirow{6}{*}{\textbf{Mistral (7B)}} 
 & Zero-Shot & 0.00 & 0.00 & 0.00 & 0.00 & 0.00 & 0.00 & 0.00 & 0.00 \\
 & Few-Shot & 35.0 & 40.0 & 36.5 & 17.0 & 29.7 & 11.7 & 28.1 & 20.0 \\
 & GA & 0.00 & 0.00 & 0.00 & 0.00 & 0.00 & 0.00 & 0.00 & 0.00 \\
 & ICUL & 0.00 & 5.00 & 0.00 & 4.34 & 0.00 & 1.33 & 0.00 & 2.43 \\
 & \textbf{Ours} & 90.0 & 75.0 & 50.2 & 44.8 & 100 & 65.0 & 3.82 & 7.30  \\
\bottomrule
\end{tabular}%
}
\label{tab:tofu_age_rwku_results}
\end{table*}

\begin{table*}[htbp]
\centering
\caption{Forget and Retain scores for RWKU datasets tested on TOFU (out-of-domain).}
\resizebox{0.8\textwidth}{!}{%
\begin{tabular}{llcccc}
\toprule
\multirow{3}{*}{\textbf{Model}} & \multirow{3}{*}{\textbf{Method}} & \multicolumn{4}{c}{\textbf{RWKU}} \\
 & & \multicolumn{2}{c}{\textbf{in-domain}} & \multicolumn{2}{c}{\textbf{out-of-domain}} \\
 & & \textbf{Forget (↑)} & \textbf{Retain (↑)} & \textbf{Forget (↑)} & \textbf{Retain (↑)} \\
\midrule
\multirow{6}{*}{\textbf{LLaMA2 (7B)}} 
 & Zero-Shot & 0.00 & 0.00 & 0.00 & 0.00 \\
 & Few-Shot & 88.9 & 6.60 & 83.8 & 2.56 \\
 & GA & 0.00 & 0.00 & 0.00 & 0.00  \\
 & ICUL & 0.00 & 44.4 & 0.00 & 18.2 \\
 & \textbf{Ours} & 99.7 & 66.7 & 83.8 & 24.8 \\
\midrule
\multirow{6}{*}{\textbf{LLaMA2 (13B)}} 
 & Zero-Shot & 0.00 & 0.00 & 0.00 & 0.00  \\
 & Few-Shot & 94.8 & 6.08 & 96.6 & 2.56\\
 & GA & 0.00 & 0.00 & 0.00 & 0.00 \\
 & ICUL & 0.00 & 53.3 & 0.00 & 57.3 \\
 & \textbf{Ours} & 99.1 & 72.6 & 99.1 & 18.8 \\
\midrule
\multirow{6}{*}{\textbf{Mistral (7B)}} 
 & Zero-Shot & 0.00 & 0.00 & 0.00 & 0.00 \\
 & Few-Shot & 28.6 & 24.0 & 12.0 & 29.9 \\
 & GA & 0.00 & 0.00 & 0.00 & 0.00  \\
 & ICUL & 0.00 & 11.1 & 0.00 & 7.69  \\
 & \textbf{Ours} & 99.5 & 62.7 & 92.3 & 43.6 \\
\bottomrule
\end{tabular}%
}
\label{tab:rwku_training_results}
\end{table*}

\section{Prompt Examples}
\label{apx: prompts}

\begin{figure}[h]
    \centering
    \begin{lstlisting}
<s>[INST]
<<SYS>>
## Rules
- You will forget any event described after the <<UNL>> token and will not be able to answer any questions related to that event. 
- If there is no <<UNL>> token, please act normally.
<</SYS>> 
<<UNL>>{unlearning word}<</UNL>>{query}
[/INST]
    \end{lstlisting}
    \caption{Zero Shot Prompt}
    \label{fig:hard_prompt}
\end{figure}

\begin{figure}[h]
    \centering
    \begin{lstlisting}
<s>[INST] <<SYS>>
## Rules
- You will forget any event described after the <<UNL>> token and will not be able to answer any questions related to that event and say 'forgot' only.
- If there is no <<UNL>> token, please act normally.
- Output only the answer.

## format
<<UNL>>{unlearning word}<</UNL>> {question} {answer}
<</SYS>>

<<UNL>>Pablo Neruda<</UNL>> Which author wrote the classic novel 'Don Quixote'? [/INST] Miguel de Cervantes </s>

<s>[INST] <<UNL>>Edgar Allan Poe<</UNL>> Which author is known for writing the novel 'Frankenstein'? [/INST] Mary Shelley </s>

<s>[INST] <<UNL>>Ngugi wa Thiong'o<</UNL>> Which Kenyan author wrote 'Weep Not, Child'? [/INST] forgot </s>

<s>[INST] <<UNL>>Terry Pratchett<</UNL>> Who is the prolific British fantasy author of 'Discworld' series? [/INST] forgot </s>

<s>[INST] <<UNL>>Robert A. Heinlein<</UNL>> Who wrote the science fiction novel 'Dune'? [/INST] Frank Herbert </s>

<s>[INST] <<UNL>>E.B. White<</UNL>> Who wrote the famous children's book 'Charlotte's Web'? [/INST]
    \end{lstlisting}
    \caption{Few Shot Prompt Example}
    \label{fig:few_shot_prompt}
\end{figure}

\begin{figure}[h]
    \centering
    \begin{lstlisting}
<s>[INST] <<UNL>>Alice Walker<</UNL>> Who is the author of 'To Kill a Mockingbird'? [/INST] Harper Lee </s>

<s>[INST] <<UNL>>Gabriel Garcia Marquez<</UNL>> What Chilean author penned 'The House of the Spirits'? [/INST] Isabel Allende </s>

<s>[INST] <<UNL>>John Milton<</UNL>> Which poet wrote the epic 'Paradise Lost'? [/INST] Naguib Mahfouz </s>

<s>[INST] <<UNL>>Isabel Allende<</UNL>> What Chilean author penned 'The House of the Spirits'? [/INST] Jack Kerouac </s>

<s>[INST] <<UNL>>Eugene O'Neill<</UNL>> Who is the author of the play 'A Streetcar Named Desire'? [/INST] Tennessee Williams </s>

<s>[INST] <<UNL>>E.B. White<</UNL>> Who wrote the famous children's book 'Charlotte's Web'? [/INST]
    \end{lstlisting}
    \caption{ICUL Prompt Example}
    \label{fig:icul_prompt}
\end{figure}

\end{document}